\documentclass{article}
\usepackage{spconf,amsmath,amssymb,graphicx,booktabs,url}
\usepackage{colortbl}
\usepackage{tcolorbox}
\usepackage{xcolor}

\title{MICRO: Multi-Fidelity Active Search for Severe Error Discovery}

\name{Orlando Leone$^{1}$, Niclas Pokel$^{1,2}$,  Pehuén Moure$^{1,2}$, Yingqiang Gao$^{3}$, Roman Boehringer$^{1}$}
      
\address{$^{1}$Institute of Neuroinformatics, University of Zurich and ETH Zurich, Switzerland \\
         $^{2}$ETH AI Center, Switzerland \\
         $^{3}$Department of Computational Linguistics, University of Zurich, Switzerland \\
      \texttt{\{oleone,npokel,pehuen,roman\}@ini.ethz.ch,}
    \texttt{yingqiang.gao@cl.uzh.ch}
}

\begin{document}

\maketitle

\begin{abstract}

Human feedback can vary in cost and informativeness. Strong feedback can reveal severe errors but is costly, so cheaper quality ratings can help decide which items to annotate. We propose MICRO (\textbf{M}ulti-Fidelity \textbf{I}mpact \textbf{C}lustered \textbf{Ro}llout), an active search framework that allocates a shared budget to these feedback types to maximise confirmed severe error discoveries. MICRO jointly models ratings and annotation losses conditional on item features to steer acquisition. It clusters acquisitions by their predicted impact on severity probabilities to select diverse candidates, then uses rollout to estimate their discovery value. Experiments on WMT20 English--German show that ratings improve both loss reconstruction and severity prediction. MICRO achieves the highest mean discovery count across four budget and rating cost settings, with similar performance to adapted MF-ENS in one and significant gains over all six comparison policies, including two rollout controls, in the other three ($p<.001$). 

\end{abstract}
\section{Introduction}
\label{sec:introduction}
Human feedback can vary substantially in informativeness and effort. For example, in personalised ASR for people with complex disabilities, explicitly correcting transcription errors can be very burdensome~\cite{guldimann2024speech}. In contrast, providing a simple quality rating is considerably easier. Machine translation deployments~\cite{dewilde2026matias} present a similar asymmetry where direct quality assessment can be substantially easier and cheaper than one-to-one correction and post-editing. In particular, Graham et al.~\cite{graham2017improving} estimated that professional post-editing can cost roughly 10--20 times as much as direct assessment. 

To exploit this difference in cost and informativeness, we propose an approach that combines weak and strong feedback~\cite{zhang2015active} under a shared acquisition budget. Strong feedback reveals an annotation loss, while weak feedback provides a scalar quality rating, much like bandit feedback~\cite{kreutzer2017bandit}. 

Active search, a variant of active learning~\cite{settles2009active}, aims to discover target examples under a limited acquisition budget~\cite{garnett2012active}. Multi-fidelity approaches extend this problem by allowing observations with different costs and levels of reliability~\cite{nguyen2021mfens}. In our setting, the relationship between quality ratings and annotation loss is uncertain and must be inferred. We use this mapping for quality estimation (QE), enabling active search to identify severe errors under a limited acquisition budget.

Quality ratings do not necessarily exactly correspond to annotation loss: for instance, perceived ASR quality can differ from measures based on word error~\cite{kim2022evaluating}. Therefore, we model ratings as noisy observations related to annotation loss.

We propose MICRO (Multi-fidelity Impact Clustered Rollout), a framework for discovering severe errors using strong and weak feedback under a shared budget. MICRO jointly models annotation losses and quality ratings conditional on item features, then clusters acquisitions by their predicted impact on posterior severity probabilities to select diverse candidates for rollout~\cite{bertsekas1997rollout}.

Targeted error discovery can reveal systematic model failures more efficiently~\cite{lakkaraju2017identifying, ribeiro2020beyond}, providing examples which can then support fine-tuning and evaluation. In this work, we evaluate feedback reconstruction and active search, demonstrating that MICRO discovers more severe errors on average than the comparison policies under the same acquisition budget.

\section{Problem Formulation}
\label{sec:problem}

Item features, such as input characteristics, embeddings or model confidence scores, can provide information about the true latent (annotation) loss~\cite{specia2010estimating}. We therefore model this dependence, and then treat ratings as noisy observations related to annotation loss and these features.

Let $\mathcal{U}$ be a large pool of unlabelled items. Each segment $i$ has known features $\boldsymbol{x}_i$, unknown annotation loss $Y_i$ and unknown ratings $R_{ij}$, where $j$ indexes the segment's ratings. We define severity as $S_i = \mathbf{1}\{Y_i>\tau\}$ for a threshold $\tau = \hat{Q}_{0.90}(\{y_i:i\in \mathcal{C}\})$, where $\mathcal{C}$ is a separate calibration set of segments and $\hat{Q}_{0.90}$ is its empirical 90th percentile. Only strong feedback which reveals a loss exceeding $\tau$ confirms a severe error discovery.

An acquisition policy~\cite{garnett2012active} selects an action $a$ at step $t$ using the set of past observations $\mathcal{D}_t$. We assign annotations a cost of $1$ and ratings a cost $\rho$ with $0<\rho<1$, and limit the total acquisition cost by setting a budget $B$ shared between both feedback modes. 

Our objective is thus to choose a policy $\pi$ which maximises the expected number of confirmed (annotated) severe segments:
\begin{equation}
\max_{\pi}\;\mathbb{E}_{\pi} \left[U_B \mid \mathcal{D}_0 \right], \qquad
U_B = \sum_{i\in \mathcal{A}} \mathbf{1}\{y_i>\tau\},
\end{equation}

where $\mathcal{A}$ is the set of annotated segments. We thus define the expected immediate reward of an action as 
\begin{equation}\bar{r}_t(a) = \begin{cases} P(Y_i > \tau \mid \mathcal D_t), & a = \operatorname{annotation}(i),\\
0, & a = \operatorname{rating}(i).
\end{cases}\end{equation}

Since utility depends on labels, observations can inform later acquisitions. A policy must therefore balance annotations' immediate discoveries against the information value of both feedback types for selecting better annotations.

\section{MICRO}
\label{sec:micro}

\subsection{Feedback Reconstruction}
\label{ssec:feedback}

We use a joint Gaussian model~\cite{rasmussen2006gaussian} for feedback reconstruction and later active search. This provides the joint predictive distribution of both annotation losses and ratings, $p(Y,R\mid \boldsymbol{x})$:
\begin{equation}
\begin{bmatrix}Y_i\\R_{ij}\end{bmatrix}
=
\begin{bmatrix}
\boldsymbol{x}_i^\top\boldsymbol{w}_Y\\
\boldsymbol{x}_i^\top\boldsymbol{w}_R
\end{bmatrix}
+\boldsymbol s_i+\boldsymbol h_{d(i)}
+\begin{bmatrix}0\\e_{ij}\end{bmatrix},
\end{equation}

where $\boldsymbol{w}_Y$ and $\boldsymbol{w}_R$ are learned weights with a Gaussian posterior, and $e_{ij}\sim\mathcal N(0,\sigma_R^2)$ is independent rating noise. 
The vectors $\boldsymbol s_i$ and $\boldsymbol h_{d(i)}$ contain loss and rating offsets for segment $i$ and its document--system group $d(i)$:
\[
\boldsymbol s_i\sim\mathcal N(\boldsymbol{0},\boldsymbol\Sigma_\mathrm{seg}),\qquad
\boldsymbol h_d\sim\mathcal N(\boldsymbol{0},\boldsymbol\Sigma_\mathrm{group}).
\]

Correlated offsets let ratings inform losses within and across segments in the same document--system group.

\subsection{Multi-Fidelity Active Search}
\label{ssec:mf-as}

Let $O_a$ be the observation returned by $a$ (which can be either an annotation or a rating), and $\boldsymbol Z$ the vector of latent losses. Given $\mathcal D_t$, these losses have mean
$\boldsymbol\mu_t$ and covariance $\boldsymbol\Sigma_t$. We then define the following posterior moments:
\[
\begin{aligned}
m_{a,t}&=\mathbb E[O_a\mid\mathcal D_t],&
v_{a,t}&=\operatorname{Var}(O_a\mid\mathcal D_t),\\
\boldsymbol c_{a,t}
&=\operatorname{Cov}(\boldsymbol Z,O_a\mid\mathcal D_t).
\end{aligned}
\]

Under the joint Gaussian model, with fitted variance parameters held fixed, observing $O_a = o_a$ gives the standard Gaussian posterior updates~\cite{frazier2009knowledge}
\begin{equation}\begin{aligned}\boldsymbol\mu_{t+1} &= \boldsymbol\mu_{t} + \frac{\boldsymbol c_{a,t}}{v_{a,t}}(o_a-m_{a,t}), \\
\boldsymbol\Sigma_{t+1} &= \boldsymbol\Sigma_{t} - \frac{\boldsymbol c_{a,t} \boldsymbol c_{a,t}^\top}{v_{a,t}}.\label{eq:posterior-update}\end{aligned}\end{equation}

Thus for an unannotated segment $i$, the posterior severity probability (with standard normal CDF $\Phi$) is 
\begin{equation}
    p_{t,i} = \Phi \left(\frac{\mu_{t,i}-\tau}{\sqrt{\Sigma_{t,ii}}}\right).
\end{equation}
Holding \(\Sigma_{t,ii}\) fixed, using a first-order Taylor approximation with respect to $\mu_{t,i}$ and then substituting from \eqref{eq:posterior-update}:
    \[\begin{aligned}
    \Delta p_{t,i} &\approx \frac{\Phi'((\mu_{t,i}-\tau)/\sqrt{\Sigma_{t,ii}})}{\sqrt{\Sigma_{t,ii}}}\Delta\mu_{t,i} \\
    &= \frac{\Phi'((\mu_{t,i}-\tau)/\sqrt{\Sigma_{t,ii}})}{\sqrt{\Sigma_{t,ii}}} \frac{ c_{a,t,i}}{v_{a,t}}(O_a-m_{a,t}).
    \end{aligned}
\]

Following the expected model output change principle~\cite{freytag2014selecting}, we define an impact vector that approximates how an observation influences each segment's severity probability:
\begin{equation}
    \psi_{a,t,i} = \frac{\Phi'((\mu_{t,i}-\tau)/\sqrt{\Sigma_{t,ii}})}{\sqrt{\Sigma_{t,ii}}} \frac{ c_{a,t,i}}{\sqrt{v_{a,t}}}.
\end{equation}

Under this approximation, we obtain the expected impact 
\begin{equation}\begin{aligned}
\mathbb E\left[\|\Delta\boldsymbol p_t\|_1 \mid\mathcal D_t\right] &\approx 
    \mathbb{E}\left[ \|\boldsymbol{\psi}_{a,t} \|_1 \left|\frac{O_a-m_{a,t}}{\sqrt{v_{a,t}}} \right| \mid \mathcal D_t\right]\\
    &= \|\boldsymbol{\psi}_{a,t}\|_1\sqrt{\frac{2}{\pi}},
\end{aligned}\end{equation}

where $\sqrt{2/\pi}$ is the mean absolute value of a standard normal variable. For an annotation of $i$, we set $\psi_{a,t,i} = 0$ (self-impact coordinate), since we score immediate discovery by $\bar{r}_t(a)$. Hence, MICRO works as follows:
\begin{enumerate}
\item We cluster possible annotations and ratings separately into $K/2$ clusters each, running $k$-means~\cite{lloyd1982least} on their impact vectors $\boldsymbol{\psi}_{a,t}$. We then score annotations by $\bar{r}_t(a)$ and ratings by $\|\boldsymbol{\psi}_{a,t}\|_1$ ($\sqrt{2/\pi}$ does not affect rankings). To reduce redundancy among rollout candidates, we construct a root pool of size $K$ by selecting the highest scoring candidate per cluster: diversity-based selection also appears in active learning~\cite{ash2020deep}. 

\item We evaluate each root over $M$ simulated trajectories. For each trajectory, we sample the root's observation to update the posterior, then simulate greedy acquisitions based on $\bar{r}_t(a)$ from the full annotation pool, conditioning on every acquisition before choosing the next. To score roots at residual budget $\delta<1$, we assign fractional reward $\delta p$ to the next greedy annotation with severity probability $p$. Greedy trajectories only use annotations because ratings have $\bar{r}_t(a) = 0$. While full look-ahead is computationally intractable, our rollout evaluates $K$ roots using $M$ simulated trajectories of at most $H$ sequential steps, thus requiring only $O(KMH)$ simulated steps. 

\item We acquire the root with the greatest estimated reward averaged over all the simulated trajectories. We then observe its feedback, update the posterior and then subtract its cost from $B$. 

\item We repeat steps 1 through 3 until less than one annotation's cost is left in the budget. A rating can be acquired only if it leaves budget for an annotation to follow.
\end{enumerate}

\section{Experimental Setup}
\label{sec:experimental}

\subsection{Feedback Reconstruction}
\label{ssec:feedback-setup}

We benchmark the model on English--German from the 2020 Workshop on Machine Translation (WMT20) dataset~\cite{freitag2021experts}. Translations are scored with strong Multidimensional Quality Metrics (MQM) and a weak 0--6 professional Scalar Quality Metric (pSQM): there are three pSQM ratings per segment. 

We choose general purpose text features, concatenating frozen LaBSE embeddings~\cite{feng2022language} of source and translation texts, then applying PCA to 32 dimensions and standardising each component. We use document-disjoint calibration, development and test splits. PCA, model fitting and standardisation use calibration data only. 

For reconstruction, we also optionally calibrate the posterior mean using two monotone functions:
\[
\widetilde y_i = c_Y(\widehat{y}_i),
\qquad
\widetilde p_i = c_S(\widehat{y}_i),
\]
where $c_Y$ and $c_S$ map the posterior mean to MQM loss and $P(Y_i>\tau)$, respectively. Both functions are fitted with isotonic regression on calibration data. We avoid selection bias from the acquisition policy by observing feedback at random.

\subsection{Multi-Fidelity Active Search}
\label{ssec:mf-as-setup}

We evaluate active search by confirmed severe errors discovered, over 1024 paired held-out English--German pools per setting at rating costs $\rho \in \{1/16, 1/8\}$ and budgets $B \in \{16, 32\}$. These costs represent hypothetical scenarios, not measured MQM/pSQM costs. Each pool contains 128 segments from the test set, which are then shared across policies.

Besides the standard greedy baseline, we also benchmark against annotation-only ENS~\cite{jiang2017ens}, which evaluates every annotation by its immediate reward and the expected top-probability batch value for the remaining budget. We adapt ENS to numerical annotation losses and estimate its expected utility by using 128 samples from the Gaussian predictive distribution. We adapt MF-ENS~\cite{nguyen2021mfens} from parallel queries to feedback under a shared budget (and numerical annotation losses). It scores each candidate by simulating its observation, followed by a batch of ratings and then a greedy batch of annotations. For each candidate, we simulate 16 possible outcomes and use 32 simulations each for batch size selection and independent evaluation. In line with MF-ENS's exploration window of one strong-query duration, we limit each simulated rating batch to one annotation's cost. Simulated batch ratings update only their own segment, while actual observations update the full posterior. ``Screen-then-annotate'' acquires only the highest scoring ratings, until it reaches a screening quota of $B/(4\rho)$ and then annotates greedily. We then include two rollout controls: ``undiversified rollout'' selects candidates by highest reward/impact without clustering's explicit diversity objective, whereas ``block-based rollout'' enforces diversity by selecting a candidate per document--system block (instead of per cluster). All policies share the fitted feedback model. 

While adapted MF-ENS evaluates all legal acquisitions, MICRO only evaluates $K=12$ candidates with $M = 128$ trajectories, chosen for computational practicality (the calibration mentioned in Section~\ref{ssec:feedback-setup} is not applied in active search). We report mean discoveries and paired differences with 95\% confidence intervals and unadjusted two-sided paired $t$-test $p$-values.

\section{Results and Discussion}
\label{sec:results}

\subsection{Feedback Reconstruction}
\label{ssec:feedback-results}
\begin{table}[htbp]
\centering
\small
\caption{Held-out reconstruction of MQM loss from translation features and pSQM ratings on English--German. Metrics are averaged within source documents and across documents. Rating-only
baselines use calibration-fitted isotonic regression.}
\label{tab:reconstruction}
\setlength{\tabcolsep}{3pt}
\begin{tabular}{@{}lcrr@{}}
\toprule
Method & Ratings & MAE $\downarrow$ & Brier $\downarrow$ \\
\midrule
Pre-rating joint model & $0/3$ & $1.762$ & $.1248$ \\
Rating-only model      & $1/3$ & $1.481$ & $.1029$ \\
Joint model            & $1/3$ & $1.461$ & $\mathbf{.0994}$ \\
Calibrated joint model & $1/3$ & $\mathbf{1.430}$ & $\mathbf{.0994}$ \\
Rating-only model      & $2/3$ & $1.360$ & $.0929$ \\
Joint model            & $2/3$ & $1.366$ & $.0897$ \\
Calibrated joint model & $2/3$ & $\mathbf{1.316}$ & $\mathbf{.0892}$ \\
Rating-only model      & $3/3$ & $1.302$ & $.0876$ \\
Joint model            & $3/3$ & $1.316$ & $.0847$ \\
Calibrated joint model & $3/3$ & $\mathbf{1.259}$ & $\mathbf{.0844}$ \\
\bottomrule
\end{tabular}
\end{table}

Table~\ref{tab:reconstruction} shows more ratings improve loss reconstruction (and in turn severity prediction). Comparing the calibrated three rating model to pre-rating predictions, MAE is reduced from 1.762 to 1.259 and Brier score from .1248 to .0844. Although the joint model's MAE is slightly higher than the rating-only baseline with two or three ratings, it consistently improves Brier score. The calibrated joint model reaches the lowest MAE and Brier scores at every rating count.
\subsection{Multi-Fidelity Active Search}
\label{ssec:mf-as-results}

\begin{table}[htbp]
\centering
\small
\caption{Severe errors discovered over 1024 paired pools per setting. Differences, confidence intervals and $p$-values compare each method with MICRO (method minus MICRO).}
\label{tab:mf-as}
\setlength{\tabcolsep}{3pt}
\begin{tabular}{@{}lrrrr@{}}
\toprule
Method & Mean & Diff. & 95\% CI & $p$ \\
\midrule
\multicolumn{5}{@{}c@{}}{$\rho=1/16,\quad B=16$} \\
\addlinespace[2pt]
Greedy                & $3.870$ & $-1.338$ & $[-1.435,-1.241]$ & $<.001$ \\
ENS                   & $3.871$ & $-1.337$ & $[-1.433,-1.241]$ & $<.001$ \\
Adapted MF-ENS        & $5.032$ & $-0.176$ & $[-0.237,-0.114]$ & $<.001$ \\
Screen-then-annotate  & $4.731$ & $-0.477$ & $[-0.550,-0.403]$ & $<.001$ \\
Undiversified rollout & $5.078$ & $-0.130$ & $[-0.184,-0.076]$ & $<.001$ \\
Block-based rollout   & $5.084$ & $-0.124$ & $[-0.177,-0.072]$ & $<.001$ \\
MICRO                 & $\mathbf{5.208}$ & -- & -- & -- \\
\midrule
\multicolumn{5}{@{}c@{}}{$\rho=1/16,\quad B=32$} \\
\addlinespace[2pt]
Greedy                & $6.502$ & $-3.471$ & $[-3.597,-3.345]$ & $<.001$ \\
ENS                   & $6.512$ & $-3.461$ & $[-3.586,-3.336]$ & $<.001$ \\
Adapted MF-ENS        & $9.500$ & $-0.473$ & $[-0.541,-0.405]$ & $<.001$ \\
Screen-then-annotate  & $8.874$ & $-1.099$ & $[-1.195,-1.003]$ & $<.001$ \\
Undiversified rollout & $9.727$ & $-0.246$ & $[-0.303,-0.189]$ & $<.001$ \\
Block-based rollout   & $9.761$ & $-0.212$ & $[-0.267,-0.156]$ & $<.001$ \\
MICRO                 & $\mathbf{9.973}$ & -- & -- & -- \\
\midrule
\multicolumn{5}{@{}c@{}}{$\rho=1/8,\quad B=16$} \\
\addlinespace[2pt]
Greedy                & $3.828$ & $-0.437$ & $[-0.523,-0.350]$ & $<.001$ \\
ENS                   & $3.830$ & $-0.435$ & $[-0.522,-0.347]$ & $<.001$ \\
Adapted MF-ENS        & $4.248$ & $-0.017$ & $[-0.070,0.037]$  & $.543$ \\
Screen-then-annotate  & $4.103$ & $-0.162$ & $[-0.212,-0.112]$ & $<.001$ \\
Undiversified rollout & $4.220$ & $-0.045$ & $[-0.093,0.003]$  & $.069$ \\
Block-based rollout   & $4.236$ & $-0.028$ & $[-0.075,0.018]$  & $.232$ \\
MICRO                 & $\mathbf{4.265}$ & -- & -- & -- \\
\midrule
\multicolumn{5}{@{}c@{}}{$\rho=1/8,\quad B=32$} \\
\addlinespace[2pt]
Greedy                & $6.695$ & $-1.522$ & $[-1.641,-1.404]$ & $<.001$ \\
ENS                   & $6.695$ & $-1.522$ & $[-1.643,-1.402]$ & $<.001$ \\
Adapted MF-ENS        & $7.962$ & $-0.256$ & $[-0.329,-0.182]$ & $<.001$ \\
Screen-then-annotate  & $7.876$ & $-0.342$ & $[-0.411,-0.273]$ & $<.001$ \\
Undiversified rollout & $8.032$ & $-0.186$ & $[-0.242,-0.129]$ & $<.001$ \\
Block-based rollout   & $8.029$ & $-0.188$ & $[-0.245,-0.132]$ & $<.001$ \\
MICRO                 & $\mathbf{8.218}$ & -- & -- & -- \\
\bottomrule
\end{tabular}
\end{table}
MICRO discovers the largest mean number of severe errors in all four settings (Table~\ref{tab:mf-as}), consistently outperforming greedy, ENS and screen-then-annotate ($p<.001$). The two annotation-only policies, greedy and ENS, achieve similarly low mean yields.

Its relative advantage over these baselines increases with larger budgets. At $\rho=1/16$, the improvements rise from $34.6\%$ to $53.4\%$ over greedy, from $34.5\%$ to $53.1\%$ over ENS, and from $10.1\%$ to $12.4\%$ over screen-then-annotate, as $B$ increases from $16$ to $32$. The same holds at $\rho = 1/8$, though the gains are smaller. 

Adapted MF-ENS and both rollout controls achieve higher mean yields than greedy, ENS and screen-then-annotate in all settings. At $\rho=1/8$, $B=16$, MICRO has a similar mean yield ($p=.543$, $p=.069$, $p=.232$), and significantly outperforms them at the other three settings ($p<.001$). These comparisons support impact-based clustering, and a plausible explanation for the advantage is that it can reduce overlap and redundancy among candidate roots, allowing rollout to consider a broader range of acquisitions.

\subsection{Limitations}
\label{ssec:Limitations}

Although personalised ASR motivates MICRO, we evaluate only on WMT20 English--German translations. Feedback costs are also assumed and standardised: in practice, annotation and rating costs may vary due to factors like item length, error severity or user fatigue. 

In additional experiments at $\rho\in \{1/4,1/2\}$, we did not observe any significant improvements by MICRO and the two rollout controls over the baseline acquisition policies. These results suggest that ratings' informativeness may not justify their cost in those settings, as their value is dataset-dependent.

Since rollout trajectories only include annotations, MICRO is limited in evaluating the benefit of acquiring multiple ratings jointly. Our MF-ENS comparison adapts its parallel setting and binary labels to continuous feedback and a shared budget, so other adaptations could yield different results.

MICRO's $k$-means step adds additional computational overhead, and we did not compare different methods under equal wall-clock time. However, in our runtime checks, clustering always took over an order of magnitude less time than the rollout phase.

\section{Conclusion}
\label{sec:conclusion}

Although weak feedback cannot directly confirm a discovery, it can improve quality estimation and guide the acquisition of strong feedback. MICRO jointly models annotation losses and quality ratings, performing active search by combining impact-based clustering with rollout to select acquisitions under a shared acquisition budget. 

Our experiments on WMT20 English--German show improved loss reconstruction and more severe error discoveries when ratings are sufficiently cheap. MICRO achieves the highest mean discovery count across all settings, and significantly outperforms all comparison policies in three settings, including both large budget settings. Comparisons with rollout controls support the benefit of impact-based clustering.

These findings motivate further study on multi-fidelity active search and its application specifically to personalised ASR. Future work should evaluate policies under equal computational budgets and realistic user effort, as well as account for sequences of weak feedback acquisition.


\newpage
\bibliographystyle{IEEEbib}
\bibliography{strings,refs}

\end{document}